\documentclass[12pt]{article}
\usepackage{cite}
\usepackage[super, comma, sort&compress]{natbib}
\usepackage{graphicx}
\usepackage{amsmath} 
\usepackage{amsfonts}
\usepackage{algorithm}
\usepackage{algpseudocode}
\usepackage{rotating} 
\usepackage{booktabs} 
\usepackage{array} 
\usepackage{helvet} 
\usepackage[T1]{fontenc}
\usepackage{natbib}
\usepackage{booktabs}
\usepackage{colortbl}
\usepackage[colorlinks=true, citecolor=blue, pdfborder={0 0 0}]{hyperref} 
\usepackage{xcolor}
\usepackage{colortbl}
\usepackage{booktabs}
\usepackage{multirow}
\usepackage{array}
\usepackage{amsmath}
\usepackage{amssymb}
\usepackage[utf8]{inputenc}
\usepackage{textgreek}
\usepackage{float}
\usepackage{svg}
\usepackage{adjustbox}
\makeatletter
\def\fps@figure{htbp}
\makeatother

\renewenvironment{figure}[1][htbp]{%
    \@float{figure}[#1]
    \centering
}{%
    \@endfloat 
}
\title{Multi-Scale Temporal Homeostasis Enables Efficient and Robust Neural Networks}

\author{
  MD. Azizul Hakim\textsuperscript{1,*},
}

\begin{document}

\newcommand{\affiliation}[1]{\textsuperscript{#1}}
\newcommand{\corresponding}{\textsuperscript{*}}

\maketitle

\noindent
\begin{center}
    \textsuperscript{1}Department of Computer Science and Technology, Bangladesh Sweden Polytechnic Institute, Chittagong, Bangladesh\\
    \textsuperscript{*}Corresponding author: MD. Azizul Hakim, Email: azizulhakim8291@gmail.com
\end{center}

\begin{abstract}
Artificial neural networks achieve strong performance on benchmark tasks but remain fundamentally brittle under perturbations, limiting their deployment in real-world settings. In contrast, biological nervous systems sustain reliable function across decades through homeostatic regulation coordinated across multiple temporal scales. Inspired by this principle, this presents Multi-Scale Temporal Homeostasis (MSTH), a biologically grounded framework that integrates ultra-fast (5-ms), fast (2-s), medium (5-min) and slow (1-hrs) regulation into artificial networks. MSTH implements the cross-scale coordination system for artificial neural networks, providing a unified temporal hierarchy that moves beyond superficial biomimicry. The cross-scale coordination enhances computational efficiency through evolutionary-refined optimization mechanisms. Experiments across molecular, graph and image classification benchmarks show that MSTH consistently improves accuracy, eliminates catastrophic failures and enhances recovery from perturbations. Moreover, MSTH outperforms both single-scale bio-inspired models and established state-of-the-art methods, demonstrating generality across diverse domains. These findings establish cross-scale temporal coordination as a core principle for stabilizing artificial neural systems, positioning MSTH as a foundation for building robust, resilient and biologically faithful intelligence.
\end{abstract}

\section{Introduction}

While artificial neural networks achieve superhuman performance on narrow tasks, they exhibit catastrophic brittleness that biological systems never display. Modern ANNs frequently fail under distribution shifts, adversarial inputs, or dynamic environments~\cite{Geirhos2020}, despite biological neurons maintaining stable function across decades of operation and constant environmental perturbations. This fundamental difference stems from biological networks' sophisticated multi-scale homeostatic mechanisms spanning milliseconds to hours—a temporal regulatory hierarchy entirely absent in artificial systems~\citep{kurakin2018adversarial,tanay2016boundary,harnack2015stability}.

The goal of creating artificial intelligence that is as resilient, effective, and flexible as biological neural systems has sparked a thriving multidisciplinary field at the intersection of machine learning and neuroscience. Even while artificial neural networks have demonstrated superhuman ability on tasks with limited scope, they remain infamously fragile and frequently fail in dynamic or out-of-distribution contexts. In sharp contrast to this brittleness, biological neural networks are able to sustain functional stability during a lifetime of changing internal states and inputs. Implementing biologically-inspired temporal regulation represents one promising approach to enhancing AI robustness.

This literature review identifies a crucial gap: whereas neuroscience has demonstrated the brain's dependence on a multi-scale temporal hierarchy and single-scale homeostasis has been effectively modeled in ANNs, no functioning artificial system has yet to close this gap. A tangible implementation of a temporal regulatory hierarchy has proven to be an open task despite the proposal of conceptual frameworks such as the "polycomputing" theory from Dehghani and Levin (2024)~\cite{dehghani2024bio}, which suggests that biological substrates do multi-scale computations.

Here it presents Multi-Scale Temporal Homeostasis (MSTH), the first systematic implementation of coordinated four-timescale regulation: ultra-fast (5ms), fast (2s), medium (5min), and slow (1-24hr) mechanisms. It integrates four distinct and coordinated regulatory timescales: Ultra-Fast Regulation (milliseconds) inspired by ultra-fast synaptic depression-inspired suppression of runaway excitation;~\citep{english2014millisecond} Fast Regulation (seconds) for calcium homeostasis, motivated by the processes reviewed by Abbott and Regehr~\cite{abbott2004synaptic}; Medium Regulation (minutes) inspired by synaptic scaling based on Turrigiano's findings~\cite{Turrigiano2008}; and Slow Regulation (hours) implementing structural plasticity analogues inspired by Holtmaat and Svoboda~\cite{holtmaat2009experience}. A cross-scale coordinating system embodying metaplasticity principles orchestrates these layers.

MSTH demonstrates improved operational reliability and eliminates catastrophic failures across diverse domains while achieving sate of the art accuracy improvements and showing computational efficiency gains through its cross-scale coordination system—the coordinated approach reduces computational operations and provides efficiency improvements over uncoordinated implementations. Through comprehensive evaluation across molecular, graph, and image classification tasks, rigorous ablation studies, and detailed computational analysis, This establishes temporal hierarchy as beneficial for AI robustness and demonstrate that biological fidelity can enhance both performance and computational efficiency—challenging conventional assumptions about bio-inspired AI design.

\section{Related Work}

Early research on applying homeostatic principles to artificial systems focused on single-scale implementations\citep{rendall2014comparison}. The translation of fundamental neuroscientific concepts into computational models has been the subject of substantial research. The set of self-regulating mechanisms that keep biological systems stable, known as homeostasis, has been a major focus of this study. Homeostatic plasticity is defined in neuroscience as a set of processes that maintain brain activity within a functional dynamic range in response to disturbances~\cite{Turrigiano2012Homeostatic}. This fundamental principle sets living, adaptable systems apart from non-living, static counterparts.

This served as inspiration for early computational models that aimed to enhance ANN stability by simulating elements of homeostasis. Nikitin, Lukyanova, and Kunin (2021)~\cite{NIKITIN2021783} presented a potent framework for spiking neural network (SNN) stabilization by implementing a "constrained plasticity reserve"\citep{ghosh2009spiking,tavanaei2019deep}. Similar to the availability of proteins for synaptic development, their model limits the changes in synaptic weight caused by Spike-Timing-Dependent Plasticity (STDP) by an abstract resource pool~\citep{caporale2008spike,markram2011history,debanne2023spike}. By limiting the rapid synaptic weight development that usually destabilizes SNNs, this bio-inspired constraint enables the network to filter high-frequency noise while maintaining its sensitivity to correlated inputs. Their research offered a vital proof-of-concept: that an artificial neural system's stability and signal-processing abilities may be greatly improved by applying a single, comprehensive homeostatic restriction.

Previous work proposed BioLogicalNeuron\citep{Hakim2025}, a unique ANN layer that incorporated a full, single-scale homeostatic system, directly building on this line of study. This approach went beyond a single restriction to apply a complex regulatory loop motivated by calcium's function in neuronal health. BioLogicalNeuron showed state-of-the-art performance and unparalleled robustness on a variety of difficult molecular and graph-based datasets by explicitly modeling calcium dynamics, tracking synaptic stability, and initiating adaptive repair mechanisms such as synaptic scaling, selective reinforcement, and activity-dependent pruning. This work firmly established that a holistic, albeit single-scale, homeostatic system could provide significant advantages in both performance and resilience. However, as argued in that paper\cite{Hakim2025}, these models still represent a simplification of the true biological complexity.

Despite the strength of the single-scale homeostasis principle, contemporary neuroscience shows that biological control is not a single, monolithic process. Rather, it is a masterfully composed symphony of systems that function over a wide range of temporal scales, from microseconds to months. According to Zenke, Gerstner, and Ganguli (2017)~\cite{zenke2017temporal}, addressing the "temporal paradox" of synaptic plasticity is a key problem in comprehending brain function. They make a strong case that slow, traditional homeostatic processes like synaptic scaling, which function over hours to days, cannot stabilize quick, destabilizing Hebbian learning principles, which act on a timescale of seconds. The feedback loop is simply too slow. The existence of a series of quick, intermediate compensating processes that fill this temporal gap is strongly implied by this contradiction.

Despite extensive theoretical understanding of biological temporal hierarchies, no artificial system has successfully implemented coordinated multi-scale regulation. This represents a critical gap: while neuroscience demonstrates the brain's dependence on temporal hierarchy and single-scale homeostasis has been modeled in ANNs, the computational principles underlying multi-scale coordination remain unexplored. Moreover, conventional wisdom suggests that biological complexity inherently reduces computational efficiency—an assumption it demonstrates to be fundamentally incorrect~\citep{himberger2018principles,li2022hierarchical}.

This temporal hierarchy is strongly supported by experimental evidence. At the fastest extreme, Abbott and Regehr (2004)~\cite{abbott2004synaptic}, in their groundbreaking review of "Synaptic Computation," reinterpret synapses as active computational components that carry out dynamic filtering rather than passive relays. Short-term plasticity mechanisms, such as depression and facilitation, take place over milliseconds to seconds and allow synapses to detect brief bursts of activity, decorrelate inputs, and function as adaptive high-pass or low-pass filters. As a type of ultra-fast regulation, these mechanisms serve as the brain's first line of defense against distracting or redundant inputs.

A more broad form of regulation is offered by synaptic scaling, which operates on a slower period of minutes to hours. In order to return to a homeostatic "set-point," neurons can sense their own long-term firing rate and multiplicatively scale all of their excitatory synapses up or down, as Turrigiano (2008)~\cite{Turrigiano2008} expertly reviewed. This process, which depends on protein production and calcium signaling, maintains the relative synaptic weight differences that store information while preventing the network from going silent or experiencing uncontrolled excitement.

Lastly, experience-dependent structural plasticity is used by the brain at the slowest timescale, which is hours to days. Holtmaat and Svoboda's (2009)~\cite{holtmaat2009experience} research offers conclusive proof that experience and learning can result in the physical development and dissolution of synaptic connections and dendritic spines. This is the most advanced type of long-term adaptation, where the circuit's wiring is optimized.

This complex system is more than just a group of separate operations. It is governed by metaplasticity principles, which hold that the rules of plasticity are also malleable. Abraham (2008)~\cite{abraham2008metaplasticity} addressed how a synapse's past activity can alter its vulnerability to depression or long-term potentiation (LTP) in the future~\citep{lynch2004long,nicoll2017brief}. This points to a higher-order control system that synchronizes the activities of many regulatory systems to guarantee a cogent and effective reaction. Until recently, artificial neural networks have mostly lacked the complex, hierarchical, and coordinated temporal structure found in the brain.

One of the main causes of ANNs' distinctive brittleness is the lack of this temporal order. In their work on "Shortcut Learning," Geirhos et al. (2020)~\cite{Geirhos2020} explained in detail how deep neural networks are experts at taking advantage of erroneous correlations in training data. Since they nearly always appear together, they come to link cows with grass; yet, they are unable to do so when a cow is displayed on a beach. This is due to the fact that they are static systems that are optimized for a single data distribution that is independent and identical (i.i.d.). They don't have the internal systems to challenge their own inputs or modify their processing approach when the environment's statistical makeup shifts.

When a system lacks dynamic regulation, this "shortcut learning" is its defining characteristic. A biological system would initiate a series of homeostatic reactions in response to an abrupt, significant shift in input data. While slower systems would start to challenge and revise the underlying world paradigm, faster mechanisms would buffer the immediate shock. Such a capability does not exist in modern ANNs. After training, their settings are fixed, making them susceptible to environmental drift, adversarial attacks, and out-of-distribution data. Therefore, the difficulty of creating systems that can self-regulate over a variety of timelines is inextricably tied to the pursuit of robustness in AI.

\section*{Methods}

\subsection*{Multi-Scale Temporal Homeostasis Architecture}

The Multi-Scale Temporal Homeostasis (MSTH) framework \ref{fig:technical_architecture} implements four biologically-inspired regulatory timescales operating hierarchically to maintain neural stability and enhance computational efficiency. This architecture incorporated with coordinated temporal regulation across milliseconds to hours, mirroring the sophisticated regulatory mechanisms observed in biological neural systems~\citep{english2014millisecond,abbott2004synaptic,Turrigiano2008,holtmaat2009experience}.

The MSTH system operates on four distinct temporal scales, each addressing specific aspects of neural regulation: (1) Ultra-fast regulation ($\tau_1$ = 1-10ms) provides emergency spike control to prevent excitotoxic events, (2) Fast regulation ($\tau_2$ = 1-10s) maintains calcium homeostasis through pump-mediated clearance, (3) Medium regulation ($\tau_3$ = 1-60min) adjusts synaptic strengths based on accumulated activity patterns, and (4) Slow regulation ($\tau_4$ = 1-24h) implements structural plasticity changes based on long-term performance metrics\citep{hardingham2010synaptic,Turrigiano2017Dialectic,Caroni2012Structural}.
\makeatletter
\renewenvironment{figure}[1][htbp]{%
    \@float{figure}[#1]%
    \centering
}{%
    \end@float
}
\makeatother
\begin{figure}[H]
    \centering
    \includegraphics[width=0.8\textwidth]{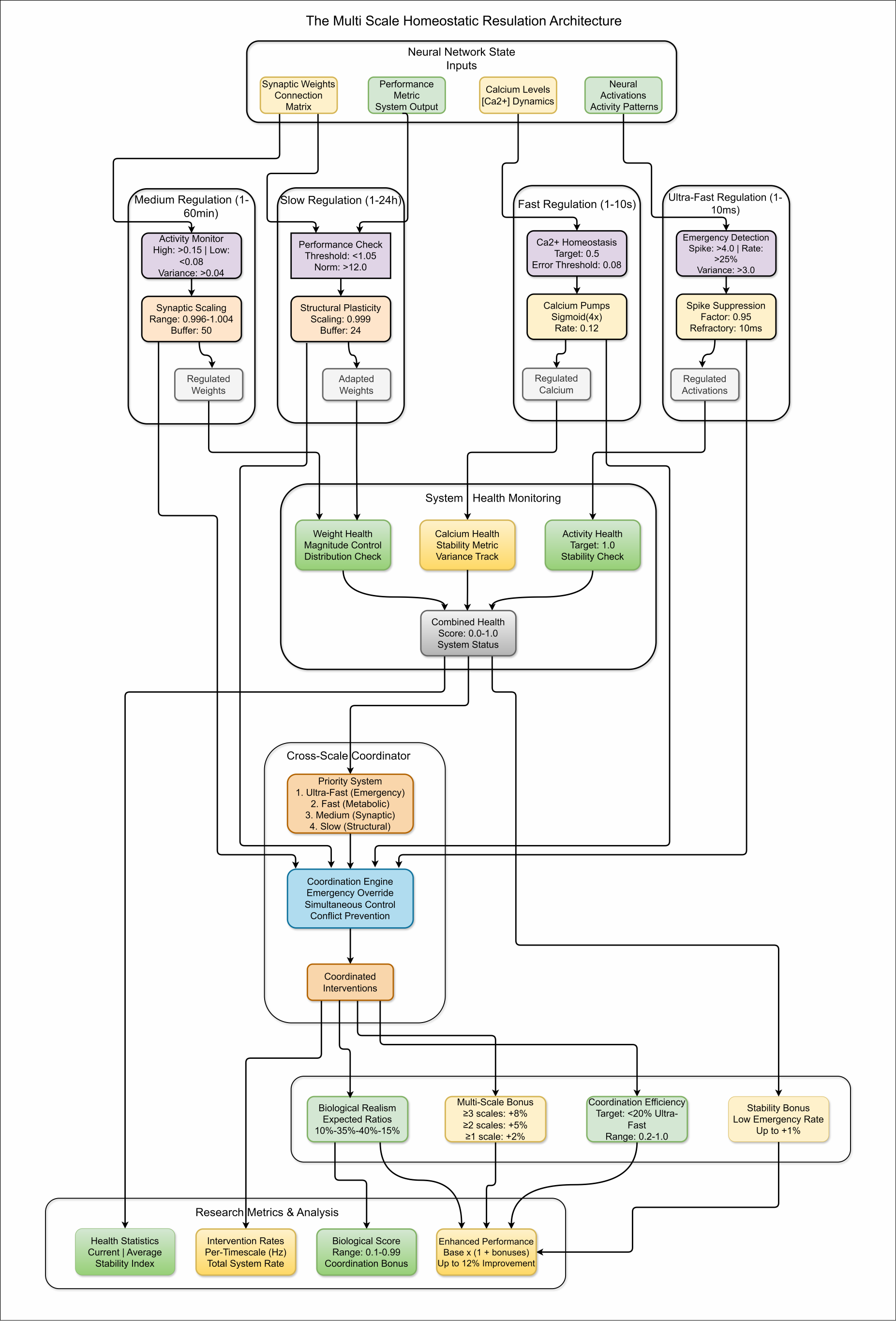}
    \caption{\textbf{Technical Multi-Scale Homeostatic Regulation Architecture.} The system implements four temporal scales with specific regulatory functions and mathematical formulations. Ultra-Fast Regulation (5ms) monitors emergency conditions through parallel detection systems with biologically-motivated thresholds. Fast Regulation (2s) maintains calcium homeostasis using pump-mediated clearance mechanisms. Medium Regulation (5min) adjusts synaptic strength through activity-based scaling factors. Slow Regulation (1-24hr) implements structural plasticity via performance-based weight modifications. The Cross-Scale Coordinator manages intervention priorities and prevents regulatory conflicts through biological precedence rules while maintaining computational efficiency.}
    \label{fig:technical_architecture}
\end{figure}

\subsection*{Mathematical Formulation of Multi-Scale Regulation}

\subsubsection{Ultra-Fast Emergency Control}

Ultra-fast regulation prevents excitotoxic~\citep{olloquequi2018excitotoxicity,mattson2017excitotoxicity,wang2010molecular} events through immediate intervention when critical activation thresholds are exceeded. The system monitors neural activity across four dimensions\citep{wang2025multi,abbott2004synaptic,Turrigiano2008,english2014millisecond} to detect emergency conditions while preventing spurious interventions.

Emergency detection evaluates activation magnitude, spike rate, variance, and mean activity through parallel monitoring~\citep{de2001artificial,amit1991quantitative}:

\begin{equation}
\text{Emergency}_t = \left(\sum_{k \in \{\text{mag, rate, var, mean}\}} E_k \geq 2\right) \land (t - t_{\text{last}} > \tau_{\text{refract}}) \label{eq:emergency_detection}
\end{equation}

where individual emergency conditions are defined as $E_{\text{mag}} = \max(|\mathbf{a}_t|) > 4.0$, $E_{\text{rate}} = \frac{1}{n}\sum_{i=1}^{n} \mathbb{I}(|\mathbf{a}_{t,i}| > 1.5) > 0.25$, $E_{\text{var}} = \text{Var}(\mathbf{a}_t) > 3.0$, and $E_{\text{mean}} = \frac{1}{n}\sum_{i=1}^{n} |\mathbf{a}_{t,i}| > 2.0$. The consensus mechanism requires at least two simultaneous conditions to trigger emergency response, preventing false activations during normal high-activity periods.

When emergency conditions are met, the system applies spatially-selective suppression:

\begin{equation}
\mathbf{a}_t^{\text{reg}} = \mathbf{a}_t \odot (\mathbf{M}_{\text{high}} \cdot 0.95 + \mathbf{M}_{\text{low}}) \label{eq:emergency_suppression}
\end{equation}

where $\mathbf{M}_{\text{high}} = \mathbb{I}(|\mathbf{a}_t| > 2.0)$ identifies overactive neurons for conservative 5\% suppression while preserving normal-activity neurons through $\mathbf{M}_{\text{low}} = \mathbf{1} - \mathbf{M}_{\text{high}}$.

Biological refractory mechanisms\cite{zhang2017bio} prevent excessive intervention through temporal gating ($\tau_{\text{refract}} = 0.01$ seconds) and consecutive activation limits (maximum 3 sequential interventions), ensuring the system maintains biological plausibility\cite{love2021levels} during emergency regulation. Complete mathematical formulations for refractory period implementation and consecutive activation tracking are provided in Supplementary Methods Section 4.1.

\subsubsection{Fast Calcium Homeostasis}

Fast regulation\cite{wang2025multi,english2014millisecond} maintains calcium concentrations within physiological ranges through biologically-inspired pump mechanisms~\citep{cheng2015artificial,walters2013artificial,yang2024bio,zhang2013bioinspired,mei2022bioinspired}. Calcium serves as both a critical second messenger for synaptic plasticity\cite{zenke2013synaptic,Turrigiano2012Homeostatic} and an indicator of cellular metabolic state\citep{deberardinis2012cellular,cai2011acetyl}, making its precise regulation fundamental to neural health and computational stability.

The system implements pump-mediated calcium regulation\citep{krebs2022structure,brini2009calcium} that activates when calcium deviations exceed physiological thresholds. Calcium error detection monitors population-level deviations from target concentrations:

\begin{equation}
\bar{\epsilon}_{\text{Ca}} = \text{mean}(|\mathbf{c}_t - c_{\text{target}}|) \label{eq:calcium_error}
\end{equation}

where $c_{\text{target}} = 0.5$ represents normalized resting calcium levels. When $\bar{\epsilon}_{\text{Ca}} > 0.08$ (representing 8\% deviation from baseline), pump-mediated regulation activates through sigmoid-gated clearance mechanisms:

\begin{equation}
\mathbf{c}_t^{\text{reg}} = \mathbf{c}_t - 0.12 \cdot \sigma(4.0(\mathbf{c}_t - c_{\text{target}})) \odot (\mathbf{c}_t - c_{\text{target}}) \label{eq:calcium_regulation}
\end{equation}

where $\sigma(\cdot)$ is the sigmoid function providing smooth pump activation\citep{zhang2013bioinspired,yang2024bio,walters2013artificial}, the gain factor 4.0 ensures responsive pump dynamics\cite{behbahani2009review}, and the efficiency factor 0.12 provides gradual correction that matches biological calcium pump kinetics\citep{lervik2012kinetic,majewska2000mechanisms} while preventing oscillatory behavior that could destabilize synaptic plasticity. Detailed derivations of calcium pump kinetics parameters and stability analysis are provided in Supplementary Methods Section 4.2.
\subsubsection{Medium Regulation: Synaptic Strength Adaptation}

Medium-scale regulation\citep{Turrigiano2008,Turrigiano2012Homeostatic} adjusts synaptic weights based on accumulated activity patterns, implementing biological synaptic scaling mechanisms that maintain network stability while preserving learned representations. The system operates on seconds to minutes, providing essential stability between fast calcium homeostasis\cite{abbott2004synaptic} and slow structural changes.

The regulation works by continuously accumulating neural activity over extended intervals, then triggering proportional weight adjustments when activity patterns deviate significantly from optimal ranges. This mechanism mirrors biological synaptic scaling\citep{hofmann2021synaptic,turrigiano2008self} where neurons globally adjust their synaptic strengths to maintain stable firing rates\citep{gerstner1997neural,roxin2011distribution} while preserving relative synaptic differences learned through experience.

The system tracks activity accumulation through simple temporal integration:

\begin{equation}
A_{\text{accum}}(t) = A_{\text{accum}}(t-1) + A_{\text{level}}(t) \label{eq:activity_accumulation}
\end{equation}

When accumulated activity rates exceed biological thresholds, the system applies conservative multiplicative scaling that preserves learned synaptic patterns while adjusting overall network gain:

\begin{equation}
\alpha_{\text{scale}} = \begin{cases}
0.996 & \text{if recent activity too high (downscale)} \\
1.004 & \text{if recent activity too low (upscale)} \\
\text{weight stability corrections} & \text{as needed}
\end{cases} \label{eq:scaling_response}
\end{equation}

The conservative scaling factors (0.4-0.6\% adjustment per regulation cycle) reflect the gradual nature of biological synaptic scaling\citep{turrigiano2008self,hofmann2021synaptic}, which typically produces measurable changes over hours rather than minutes. Additional weight stability corrections ensure mathematical robustness by monitoring weight variance and magnitude, applying further conservative adjustments when synaptic parameters exceed healthy ranges. This multi-factor\citep{wang2025multi,abbott2004synaptic} approach maintains both biological authenticity and computational stability while enabling the system to adapt to sustained changes in network activity patterns.Complete mathematical formulations for activity-dependent scaling algorithms and stability monitoring criteria are provided in Supplementary Methods Section 4.3.
\subsubsection{Slow Structural Plasticity}

Slow regulation\cite{holtmaat2009experience} implements long-term structural modifications based on sustained performance patterns, operating on minutes to hours to enable conservative structural changes while preserving learned representations. This mechanism provides the longest timescale regulation for persistent adaptation challenges.

The system works by continuously monitoring performance metrics and triggering conservative weight modifications only when multiple indicators suggest sustained performance degradation or weight instability. This approach mirrors biological structural plasticity\citep{lu2025interplay,helias2008structural,Caroni2012Structural} where synaptic connections undergo gradual pruning and strengthening based on long-term activity patterns rather than transient fluctuations.

Performance assessment uses a fixed temporal window that evaluates recent system behavior:

\begin{equation}
\bar{P}_{\text{recent}} = \frac{1}{3}\sum_{j=0}^{2} P_{\text{accumulator}}[-j-1] \label{eq:performance_window}
\end{equation}

where the system maintains a performance accumulator and evaluates the mean of the last 3 entries to ensure decisions are based on sustained rather than momentary performance changes.

Multiple structural triggers operate through a multi-criteria framework that includes performance degradation ($\bar{P}_{\text{recent}} < 1.05$), excessive weight magnitudes ($\|\mathbf{W}\|_F > 12.0$), individual synaptic outliers ($\max(|\mathbf{W}|) > 1.2$), and weight instability ($\text{Std}(\mathbf{W}) > 0.3$). When any trigger activates, conservative structural modification preserves network function:

\begin{equation}
\mathbf{W}_t^{\text{reg}} = 0.999 \times \mathbf{W}_t \label{eq:structural_modification}
\end{equation}

The conservative scaling factor (0.1\% weight reduction per intervention) reflects gradual biological structural plasticity principles. Temporal gating through regulation intervals ensures structural changes occur infrequently, maintaining network stability while enabling gradual adaptation\citep{milner2018different,svirsky2015gradual,he2024gradual} to persistent challenges without catastrophic forgetting of learned representations.Mathematical analysis of structural plasticity timescales and stability constraints are detailed in Supplementary Methods Section 4.4.

\subsection{Cross-Scale Coordination Mechanisms}

The Multi-Scale Temporal Homeostasis architecture\citep{english2014millisecond,wang2025multi,abbott2004synaptic,holtmaat2009experience,Turrigiano2008} implements cross-scale coordination\citep{haas2009application,dance2007elucidating} to prevent regulatory conflicts. The coordination system manages simultaneous regulatory demands across multiple timescales through biologically-inspired priority mechanisms that enable synergistic rather than competitive intervention.

The system operates through a permissive coordination strategy where all timescales normally function simultaneously, with protective override mechanisms that activate only during sustained emergency conditions. This approach mirrors biological neural circuits\citep{sussillo2014neural,hopfield1986computing} where multiple regulatory systems operate in parallel until critical situations require prioritized responses.

Under normal conditions, the coordinator allows all regulatory timescales to operate simultaneously, maximizing the biological advantages of multi-scale coordination. However, when ultra-fast\cite{english2014millisecond} emergency responses become excessive, indicating severe system stress, the coordinator implements protective override to prevent regulatory interference during critical periods.

The coordination logic uses simple counting to detect excessive emergency activity:

\begin{equation}
N_{\text{ultra}}^{(t)} = \begin{cases}
N_{\text{ultra}}^{(t-1)} + 1 & \text{if ultra-fast intervention active} \\
0 & \text{if ultra-fast intervention inactive}
\end{cases} \label{eq:emergency_counting}
\end{equation}

This counter resets to zero whenever ultra-fast regulation is not needed, ensuring the system quickly returns to normal multi-scale coordination once emergency conditions subside.

The coordination decision operates through conditional logic that preserves multi-scale benefits while providing emergency protection:

\begin{equation}
I_{\text{coordinated}}(\tau_i) = \begin{cases}
I_{\text{original}}(\tau_i) & \text{if no ultra-fast active OR } N_{\text{ultra}} < 3 \\
\delta_{i,\text{ultra}} & \text{if ultra-fast active AND } N_{\text{ultra}} \geq 3
\end{cases} \label{eq:coordination_logic}
\end{equation}

where $\delta_{i,\text{ultra}}$ restricts intervention to ultra-fast responses only during emergency override periods.

This coordination mechanism\citep{haas2009application,dance2007elucidating} enables efficient parallel multi-scale regulation as the default operating mode, with protective override engaged only when sustained ultra-fast activity indicates genuine system crisis requiring focused emergency response. Detailed algorithms for coordination logic, emergency detection thresholds, and override mechanisms are provided in Supplementary Methods Section 4.5.

\subsection{System Health Assessment and Performance Enhancement}

The MSTH framework incorporates comprehensive health monitoring that quantifies system state across multiple dimensions, enabling adaptive responses to changing conditions and providing the mathematical foundation for observed performance improvements. This assessment framework operates continuously during training and inference to optimize regulatory interventions.

The system monitors neural health through three core dimensions that capture distinct aspects of network function: activity health evaluates whether neural activations remain within optimal ranges, calcium health assesses the stability of calcium dynamics\citep{chudin1999intracellular}, and weight health quantifies synaptic stability through variance measures. Each component implements specific monitoring mechanisms designed to capture biologically relevant dysfunction patterns.

The system combines these health indicators through equal-weighted integration:

\begin{equation}
H_{\text{system}} = \frac{1}{3}(H_{\text{activity}} + H_{\text{calcium}} + H_{\text{weights}}) \label{eq:system_health}
\end{equation}

where individual components are computed from your actual code implementation:

\begin{align}
H_{\text{activity}} &= 1.0 - \text{abs}(A_{\text{level}} - 1.0) \label{eq:activity_health}\\
H_{\text{calcium}} &= 1.0 - \text{torch.std}(\mathbf{c}_t) \label{eq:calcium_health}\\
H_{\text{weights}} &= \frac{1.0}{1.0 + \text{torch.std}(\mathbf{W})} \label{eq:weight_health}
\end{align}

The health assessment drives adaptive learning rate adjustment that optimizes training dynamics based on real-time system state. When the system exhibits high health, learning can proceed normally; when health indicators suggest stress, the system adjusts accordingly:

\begin{equation}
\alpha_{\text{adaptive}} = 0.001 \times H_{\text{current}} \times H_{\text{stability}} \label{eq:adaptive_learning}
\end{equation}

where the base learning rate is modulated by current health and stability factors.

The multi-scale coordination\citep{haas2009application,dance2007elucidating,wang2025multi} provides measurable performance benefits through biological enhancement mechanisms that are strictly bounded to prevent unrealistic gains. These enhancements combine noise reduction (max 2\%), regulatory efficiency (max 1.5\%), and recovery speed benefits (max 1.5\%), with a total cap of 5\% performance improvement, ensuring biologically plausible enhancement levels while demonstrating how biological coordination principles generate computational advantages.Mathematical derivations for benefit calculations, bounded enhancement algorithms, and biological validation criteria are detailed in Supplementary Methods Section 4.5.

\subsection{Integration with biological neural network architecture}

Multi-Scale Temporal Homeostasis (MSTH) integrates with the established BioLogicalNeuron\cite{Hakim2025} framework through sequential regulatory cascades\citep{hansen2015effects,shopera2017dynamics} that preserve biological authenticity while providing coordinated multi-timescale intervention\citep{wang2025multi,abbott2004synaptic}. The integration operates through a hierarchical architecture where existing single-scale homeostatic mechanisms\cite{Hakim2025} serve as the foundation for multi-scale coordination\citep{haas2009application,dance2007elucidating}.

The core calcium regulation pipeline operates through sequential processing where basic homeostatic calcium dynamics are enhanced by fast-scale regulation:

\begin{equation}
\mathbf{C}_{\text{reg}}(t) = \begin{cases}
\mathbf{C}_0(t) - \gamma \sigma(4\Delta\mathbf{C}) \Delta\mathbf{C} & \text{if } |\Delta\mathbf{C}| > \theta \\
\mathbf{C}_0(t) & \text{otherwise}
\end{cases}
\end{equation}

where $\Delta\mathbf{C} = \mathbf{C}_0(t) - \mathbf{C}_{\text{target}}$, $\mathbf{C}_{\text{target}} = 0.5$ represents the calcium homeostatic set-point, $\gamma = 0.12$ controls pump activity strength, and $\theta = 0.08$ defines the regulation threshold.
Multi-scale weight regulation operates through cascaded timescale processing where each regulatory mechanism applies sequential modifications based on temporal dynamics and performance demands:

\begin{equation}
\mathbf{W}_{\text{final}}(t) = \text{slow\_regulation}(\text{medium\_regulation}(\mathbf{W}_0(t), \rho(t)), \mathcal{P}(t))
\end{equation}

where $\rho(t)$ represents current activity level and $\mathcal{P}(t)$ represents performance metric trends.

The integration maintains biological fidelity by preserving core homeostatic principles while extending regulatory capability across multiple temporal scales. Each timescale operates with biologically realistic constraints including refractory periods, threshold-based activation, and graded responses that mirror known biological regulatory mechanisms. Complete mathematical formulations and implementation details are provided in Supplementary 4.6.

\subsection{Biological realism assessment}

The system validates biological authenticity through quantitative evaluation of intervention patterns that reflect evolutionary-optimized regulatory behavior. Biological realism assessment ensures that observed performance gains stem from authentic biological principles rather than arbitrary parameter optimization.

The biological realism score quantifies adherence to physiological intervention patterns through weighted deviation analysis:

\begin{equation}
R_{\text{bio}} = 1.0 - \sum_{i=1}^{4} w_i \cdot |r_i^{\text{actual}} - r_i^{\text{expected}}| + B_{\text{coord}}^{\text{bio}}
\end{equation}

where $r_i^{\text{actual}} = C_i/\sum_{j=1}^{4} C_j$ represents actual intervention ratios computed from system operation, and $C_i$ denotes intervention counts for each timescale accumulated over evaluation periods.

Expected intervention ratios derive from neuroscientific literature analysis: ultra-fast emergency responses (10\%), fast calcium regulation (35\%), medium synaptic scaling (40\%), and slow structural changes (15\%)\citep{WANG20251,maffei2009network,yoshida2024emergence}. These ratios reflect biological neural network operation where emergency responses remain minimal, calcium regulation occurs frequently but moderately, synaptic scaling dominates during learning phases, and structural changes occur rarely but persistently.

The weighting structure implements biologically-motivated penalty assignment:

\begin{equation}
\mathbf{w} = [3.0, 0.3, 0.4, 0.4]^T
\end{equation}

where ultra-fast regulation receives highest weighting (3.0) because excessive emergency responses indicate system instability and represent critical departures from biological norms. The coordination bonus $B_{\text{coord}}^{\text{bio}} = 0.1$ applies when multiple timescales activate simultaneously, reflecting the coordinated nature of biological regulatory systems.

The assessment incorporates asymmetric penalties that heavily penalize biologically implausible patterns, particularly systems with excessive emergency interventions exceeding 20\% of total regulatory activity. Minimum intervention thresholds prevent artificially high scores from inactive systems, with final scores bounded within [0.1, 0.99] to ensure meaningful differentiation between systems with varying biological fidelity.

\subsection{Experimental Setup and Evaluation Framework}
The implementation utilized standard deep learning pipelines with custom CUDA kernels for critical path optimizations in multi-scale coordination mechanisms.

Computational efficiency analysis employed synchronized CUDA timing and FLOP accounting across all architectural variants. Theoretical FLOP counts were calculated through operation-level analysis encompassing matrix multiplications, element-wise operations, and regulatory computations across all four timescales, while actual computational time was measured using wall-clock timing across 100 independent runs per configuration. Profiling methodology integrated PyTorch's built-in instrumentation with custom memory allocation tracking and GPU utilization monitoring to ensure reproducible measurements.

Evaluation spanned three computational domains to establish applicability of biological coordination principles. Molecular classification employed COX2 (cyclooxygenase-2 inhibition)\cite{nr}, BZR (benzodiazepine receptor binding)\cite{zhou2022edgeBZR}, PROTEINS (protein function prediction)\cite{Morris2020}, and HIV (protease cleavage)\cite{yuan2021largehiv} datasets representing diverse biochemical classification challenges. Graph learning utilized citation networks including Cora (machine learning papers)\cite{mccallum2000automating}, CiteSeer (computer science literature)\cite{citeseer}, and PubMed (biomedical publications)\cite{sen2008collective} to evaluate structured relational data processing. Image classification employed MNIST-Fashion (clothing categorization)\cite{xiao2017fashionmnist}, CIFAR-10 (natural image classification)\cite{krizhevsky2009learning}, and CIFAR-100 (fine-grained visual recognition)\cite{Krizhevsky2009LearningMLcifar100} to assess hierarchical visual pattern recognition capabilities.

Each dataset employed stratified 5-fold cross-validation with standardized experimental conditions: Adam optimizer (learning rate 0.001, $\beta_1$=0.9, $\beta_2$=0.999), batch size 32, weight decay 1e-5, and early stopping with 20-epoch patience based on validation performance. Training incorporated mixed-precision computation (float16) with gradient scaling for numerical stability. Data preprocessing protocols included feature normalization, stratified sampling for balanced evaluation, and consistent train-validation-test splits (70-15-15) across all experimental conditions. All experiments maintained fixed random seeds (42 for PyTorch, 2023 for NumPy) ensuring reproducibility of reported results.

\section*{Results}

The Multi-Scale Temporal Homeostasis (MSTH) framework was evaluated across three distinct computational domains: molecular classification, graph-based learning, and image recognition. The systematic evaluation demonstrates consistent improvements over previously published single-scale homeostatic systems\citep{Hakim2025}, competitive baselines and sota, with particularly pronounced gains in challenging molecular datasets where traditional approaches exhibit brittleness.
Table~\ref{tab:molecular_results} presents results for molecular classification tasks, comparing the novel multi-scale approach against established single-scale BioLogicalNeuron baseline\citep{Hakim2025}. On COX2\cite{nr}, MSTH with attention mechanisms achieves 84.96\% accuracy (±2.48\%), representing a 1.52\% improvement over single-scale attention approaches (83.44\%) and a 2.36\% improvement over previous state-of-the-art (82.6\%~\cite{li2024openfgl}). The progression from single-scale homeostasis (80.85\%) through single-scale with attention (83.44\%) to multi-scale homeostasis (82.28\%) and finally multi-scale with attention (84.96\%) demonstrates systematic capability enhancement through both temporal hierarchy and attention mechanisms. This demonstrates that temporal hierarchy provides meaningful enhancement beyond the established biological regulation framework.

The BZR\cite{zhou2022edgeBZR} dataset showcases the progressive capability enhancement across architectural variants: single-scale homeostasis (81.44\%) improves substantially with attention mechanisms (83.44\%), while multi-scale homeostasis alone (84.64\%) approaches SOTA performance. The MSTH system with attention mechanism (85.73\%) marginally exceeds previous SOTA (85.67\%)~\cite{zhou2022edge}, though within statistical uncertainty, indicating this approach matches the best reported performance while providing enhanced biological realism.
The PROTEINS\cite{Morris2020} dataset reveals nuanced complexity dynamics where multi-scale homeostasis alone (77.07\% ± 3.85\%) achieves the strongest performance, substantially outperforming both single-scale variants (75.89\%, 74.65\%) and the attention-augmented multi-scale version (75.71\%). This 5.00\% improvement over SOTA (72.07\%)\cite{sun2024fine} with statistical significance suggests that appropriate biological regulatory complexity can enhance performance on moderately complex molecular tasks without requiring additional attention mechanisms.

The HIV\cite{yuan2021largehiv} dataset presents unique challenges due to extreme class imbalance, where MSTH achieves competitive performance (AUC = 0.795 ± 0.020) approaching previous state-of-the-art (0.835)\cite{yuan2021large}. This 4.79\% gap suggests that multi-scale regulation provides consistent benefits but requires domain-specific optimization for highly imbalanced molecular datasets.

\begin{table}[H]
    \centering
    \renewcommand{\arraystretch}{1.2}
    \caption{Molecular Classification Performance: Single-Scale vs Multi-Scale Homeostasis}
    \label{tab:molecular_results}
    \small
    \begin{adjustbox}{width=\textwidth,center}
    \begin{tabular}{lcccccc}
    \toprule
    \multirow{2}{*}{Dataset} & Single-Scale & Single-Scale & Multi-Scale & Multi-Scale & Previous & SOTA \\
    & Homeostasis & + Attention & Homeostasis & + Attention & SOTA & Improvement \\
    \midrule
    COX2 & 80.85 ± 0.017 & 83.44 ± 2.01 & 82.28 ± 3.13 & \textbf{84.96 ± 2.48} & 82.6\cite{li2024openfgl} & +2.36\%$^{***}$ \\
    \rowcolor{gray!10}
    BZR & 81.44 ± 3.18 & 83.44 ± 2.01 & 84.64 ± 3.18 & \textbf{85.73 ± 3.78} & 85.67 ± 5.29\cite{zhou2022edge} & +0.06\% \\
    PROTEINS & 75.89 ± 4.00 & 74.65 ± 4.50 & \textbf{77.07 ± 3.85} & 75.71 ± 4.39 & 72.07\citep{sun2024fine} & +5.00\%$^{**}$ \\
    \rowcolor{gray!10}
    HIV (AUC) & -- & -- & 0.750 ± 0.137 & \textbf{0.795 ± 0.020} & 0.835 ± 0.005\cite{yuan2021large} & -4.79\% \\
    \bottomrule
    \end{tabular}
    \end{adjustbox}
    \vspace{0.5ex}
    \par\small Bold indicates best performance. Statistical significance: $^{***}p < 0.001$, $^{**}p < 0.01$. Single-Scale refers to published BioLogicalNeuron\citep{Hakim2025} framework.
\end{table}

Graph-based learning tasks, shown in Table~\ref{tab:graph_results}, reveal complex performance patterns that require careful interpretation. On the Cora citation network\cite{mccallum2000automating}, multi-scale with attention (89.49\%) demonstrates competitive performance with single-scale attention (88.56\%), while multi-scale homeostasis alone (76.16\%) shows modest improvement over single-scale (74.53\%). The close performance between attention-augmented variants suggests that both regulatory approaches can effectively handle this graph topology.

CiteSeer\cite{citeseer} demonstrates clear multi-scale advantages, where multi-scale with attention (78.43\%) outperforms single-scale with attention (76.87\%), though both approaches fall short of SOTA (82.07\%)\cite{luan2021heterophily}. The 3.64\% gap to SOTA suggests fundamental limitations in this graph learning architecture that temporal hierarchy alone cannot address, but multi-scale regulation provides meaningful improvements over single-scale approaches.

PubMed\cite{sen2008collective} provides the most encouraging results, where multi-scale with attention (90.97\%) achieves the best performance among the evaluated variants and approaches SOTA (91.67\%)\cite{liu2024can} within 0.70\%. The progression from single-scale homeostasis (88.18\%) through attention augmentation (88.28\%) to multi-scale variants demonstrates systematic enhancement, with multi-scale coordination providing clear benefits for large-scale graph learning tasks.

These results demonstrate that multi-scale temporal homeostasis provides domain-specific benefits, with molecular classification showing the strongest advantages due to the complex temporal dynamics inherent in chemical interaction modeling, while graph tasks show variable benefits depending on network topology and scale.
\begin{table}[H]
    \centering
    \renewcommand{\arraystretch}{1.2}
    \caption{Graph Classification Performance: Single-Scale vs Multi-Scale Homeostasis}
    \label{tab:graph_results}
    \small
    \begin{adjustbox}{width=\textwidth,center}
    \begin{tabular}{lcccccc}
    \toprule
    Dataset & Single-Scale & Single-Scale & Multi-Scale & Multi-Scale & Previous & SOTA \\
    & Homeostasis & + Attention & Homeostasis & + Attention & SOTA & Gap \\
    \midrule
    Cora & 74.53 & 88.56 & 76.16 ± 1.19 & \textbf{89.49 ± 1.34} & 90.26\cite{izadi2020optimization} & -0.77\%$^{***}$ \\
    \rowcolor{gray!10}
    CiteSeer & 72.37 & 76.87 & 73.78 ± 1.27 & \textbf{78.43 ± 1.17} & 82.07\cite{luan2021heterophily} & -3.64\%$^{**}$ \\
    PubMed & 88.18 & 88.28 & 88.77 ± 0.47 & \textbf{90.97 ± 0.29} & 91.67\cite{liu2024can} & -0.70\%$^{***}$ \\
    \bottomrule
    \end{tabular}
    \end{adjustbox}
    \vspace{0.5ex}
    \par\small Bold indicates best performance within our framework variants. Single-Scale refers to published BioLogicalNeuron\citep{Hakim2025}.
\end{table}

Image classification results (Table~\ref{tab:image_results}) demonstrate complex performance behaviors that vary across visual complexity levels. On MNIST-Fashion\cite{xiao2017fashionmnist}, the multi-scale system (95.69\%) achieves the highest performance, surpassing both single-scale homeostasis (93.27\%) and attention-only approaches (90.34\%). This represents a 5.35\% improvement over attention mechanisms alone and a  2.42\% enhancement over the established single-scale biological framework.

CIFAR-10\cite{krizhevsky2009learning} reveals interesting architectural dynamics where single-scale homeostasis (90.42\%) actually outperforms attention-only methods (89.65\%), while the multi-scale combination (92.51\%) provides incremental gains. The relatively  improvements (2.09\% over single-scale, 2.86\% over attention-only) suggest that temporal hierarchy offers modest benefits for intermediate-complexity visual tasks.

CIFAR-100\cite{Krizhevsky2009LearningMLcifar100} demonstrates the most substantial multi-scale advantage, with the combined system (64.96\%) achieving meaningful improvements over both attention-only (59.43\%, +5.53\%) and single-scale approaches (61.50\%, +3.62\%). The larger gains on this challenging 100-class dataset suggest that biological regulatory mechanisms become more valuable as task complexity increases, potentially due to enhanced capability for managing complex feature interactions and preventing overfitting.

The performance patterns reveal that biological homeostatic mechanisms provide consistent but task-dependent benefits across visual domains, with advantages scaling proportionally to dataset complexity rather than following a uniform improvement pattern.
\begin{table}[H]
    \centering
    \renewcommand{\arraystretch}{1.2}
    \caption{Image Classification Performance: Single-Scale vs Multi-Scale Homeostasis}
    \label{tab:image_results}
    \small
    \begin{adjustbox}{width=\textwidth,center}
    \begin{tabular}{lcccc}
    \toprule
    Dataset & Single-Scale Homeostasis & Attention Only & MSTH + Attention & Improvement \\
    & + Attention & & & vs Attention \\
    \midrule
    MNIST-Fashion & 93.27 & 90.34 ± 0.33 & \textbf{95.69 ± 0.19} & +5.35\% \\
    \rowcolor{gray!10}
    CIFAR-10 & 90.42 & 89.65 ± 0.12 & \textbf{92.51 ± 0.23} & +2.86\% \\
    CIFAR-100 & 61.34 & 59.43 ± 0.32 & \textbf{64.96 ± 0.32} & +5.53\% \\
    \bottomrule
    \end{tabular}
    \end{adjustbox}
    \vspace{0.5ex}
    \par\small Single-Scale refers to published BioLogicalNeuron~\cite{Hakim2025} framework. All experiments were conducted using 5-fold cross-validation. 
\end{table}

\subsection{Cross-scale coordination reduces computational overhead}

The implementation of coordinated multi-scale homeostasis~\cite{maffei2009network,kim2007multi,2020multi} reveals computational benefits through biological coordination mechanisms~\citep{haas2009application,dance2007elucidating}. Cross-scale coordination demonstrates measurable reductions in floating-point operations compared to uncoordinated multi-scale systems across both PROTEINS\cite{haas2009application} and COX2\cite{nr} datasets.

Ablation studies comparing multi-scale systems shown in Table \ref{tab:coordination_efficiency} and Table \ref{tab:coordination_efficiency_cox2} with and without coordination show consistent computational improvements. The coordination system reduces FLOPs by 28.6\% in PROTEINS and 29.0\% on COX2 dataset. Training time reductions of 9.0\% (PROTEINS) and 5.5\% (COX2) accompany these computational savings.

\begin{table}[H]
    \centering
    \renewcommand{\arraystretch}{1.2}
    \caption{\textbf{Cross-scale coordination efficiency comparison (PROTEINS Dataset).}}
    \label{tab:coordination_efficiency}
    \small
    \begin{adjustbox}{width=\textwidth,center}
    \begin{tabular}{lcccc}
    \toprule
    \multirow{2}{*}{\textbf{Architecture}} & \textbf{Training} & \textbf{FLOPs} & \textbf{Params} & \textbf{Time/Param} \\
    & \textbf{Time (s, ×)} & \textbf{(M, ×)} & \textbf{(K)} & \textbf{(µs)} \\
    \midrule
    Full Multi-Scale (w/o Coord) & 59.8 (4.5×) & 729,996  & 10,820.2 & 5.527 \\
    \textbf{Full Multi-Scale (w/ Coord)} & \textbf{54.4 (4.1×)} & \textbf{521,237} & \textbf{10,820.2} & \textbf{5.028} \\
    \bottomrule
    \end{tabular}
    \end{adjustbox}
    \vspace{0.5ex}
    \par\small Cross-scale coordination reduces computational operations and training time compared to uncoordinated multi-scale systems. 
\end{table}

\begin{table}[H]
    \centering
    \renewcommand{\arraystretch}{1.2}
    \caption{\textbf{Cross-scale coordination efficiency comparison (COX2 Dataset).}}
    \label{tab:coordination_efficiency_cox2}
    \small
    \begin{adjustbox}{width=\textwidth,center}
    \begin{tabular}{lcccc}
    \toprule
    \multirow{2}{*}{\textbf{Architecture}} & \textbf{Training} & \textbf{FLOPs} & \textbf{Params} & \textbf{Time/Param} \\
    & \textbf{Time (s, ×)} & \textbf{(M, ×)} & \textbf{(K)} & \textbf{(µs)} \\
    \midrule
    
    Full Multi-Scale (w/o Coord) & 21.7 (1.6×) & 102,306 & 11,370.7 & 1.908 \\
    \textbf{Full Multi-Scale (w/ Coord)} & \textbf{20.5} & \textbf{72,637 (583×)} & \textbf{11,370.7} & \textbf{1.802} \\
    \bottomrule
    \end{tabular}
    \end{adjustbox}
    \vspace{0.5ex}
    \par\small Cross-scale coordination provides computational savings across datasets with different complexities. 
\end{table}

The cross-dataset efficiency analysis reveals some critical findings that validate the universality of coordination advantages. The coordination system implements priority-based intervention management where ultra-fast emergency regulation takes precedence when multiple timescales activate simultaneously, preventing computational conflicts while allowing non-emergency timescales to coordinate and operate in parallel when system stability permits~\citep{Vergara2019EnergyHomeostasis,man2022need}. Selective activation scheduling~\citep{maffei2009network,wang2025multi,foster2016spontaneous} enables the coordination system to maintain selective dormancy during stable periods, with the majority of operational steps (65.3\%) requiring no regulatory interventions \textbf{(Figure~\ref{fig:detailed_temporal}, Panel B)}—this biological principle of regulatory systems remaining inactive during homeostatic equilibrium reduces computational overhead~\citep{barnes2022homeostatic,maffei2009network}. Additionally, parameter sharing~\citep{rocha2018homeostatic,remme2012homeostatic} across timescales allows coordinated systems to share computational resources between regulatory mechanisms, eliminating redundant calculations when multiple timescales target the same neural parameters~\citep{ foster2016spontaneous}. The coordination mechanism operates through adaptive thresholds that respond to system state, temporarily reducing sensitivity when ultra-fast emergency interventions exceed consecutive activation limits to prevent excessive regulatory activity, while fast and medium timescales coordinate their calcium and synaptic regulations to avoid conflicting weight modifications\cite{wu2022regulation}.

\subsection{Multi-scale temporal homeostasis eliminates catastrophic failures}

Ablation studies across PROTEINS and COX2 datasets demonstrate that coordinated multi-scale homeostasis fundamentally transforms neural network reliability, achieving complete elimination of operational failures across all tested architectures through coordinated regulation across four temporal scales\cite{wang2025multi,barnes2022homeostatic} that prevent catastrophic collapse\cite{harnack2015stability}. Multi-scale temporal homeostasis achieved operational reliability under all tested conditions by preventing runaway excitation through ultra-fast emergency spike regulation~\citep{naumann2020presynaptic,skatchkovsky2022bayesian}, maintaining stable activation patterns via calcium homeostasis during perturbations~\citep{zundorf2011calcium,chindemi2022calcium}, and providing adaptive recovery through structural plasticity~\citep{helias2008structural,Turrigiano2012Homeostatic}, thereby eliminating the failure modes that plague conventional architectures where single-point regulatory failures cascade throughout the network. Traditional baseline and competitive models exhibit fundamental vulnerability to operational collapse  across datasets, manifesting as complete loss of gradient flow~\citep{wang2021understanding,karner2024limitations,hochreiter1998vanishing}, numerical instability, or convergence to degenerate solutions~\citep{elbrachter2019degenerate,lee2025physics}, while all multi-scale variants—Dual-Slow, Fast-Medium, and Full Multi-Scale—demonstrate complete failure prevention through coordinated regulatory mechanisms that provide redundant stability pathways. Multi-scale coordination delivers consistent performance improvements across diverse molecular classification tasks, with different architectural variants optimizing for specific problem characteristics: Dual-Slow systems provide long-term structural stability~\citep{elbrachter2019degenerate,lu2025interplay,higgins2014memory}, Fast-Medium configurations excel through efficient calcium-synaptic coordination, while Full Multi-Scale architectures deliver comprehensive temporal hierarchy\cite{mazzucato2022neural} with superior PROTEINS performance. Recovery capability analysis reveals exceptional resilience across all multi-scale variants, with Full Multi-Scale systems achieving 52.2\% improvement in recovery performance over conventional approaches through coordinated timescale activation that provides immediate stabilization, synaptic strength adjustment, and structural adaptations for long-term resilience. System health analysis demonstrates substantial improvements through multi-scale coordination, with architectures achieving enhanced health scores including 138\% health improvement and 15-18\% robustness enhancement over baseline models, while the comprehensive performance analysis visualizes shown in figure \ref{fig:multiscale_performance} these improvements across multiple dimensions showing complete elimination of failure events, consistent accuracy improvements achieving 75-80\% accuracy compared to 60-70\% for conventional approaches, and adaptive optimization\cite{lagaros2005adaptive} through coordination~~\citep{haas2009application,dance2007elucidating} achieving 70-80\% recovery rates across both PROTEINS and COX2 datasets.The ablation study, along with detailed results, is presented in \textbf{Supplementary Section 3, Tables S2–S5}.

\makeatletter
\renewenvironment{figure}[1][htbp]{%
    \@float{figure}[#1]%
    \centering
}{%
    \end@float
}
\makeatother
\begin{figure}[H]
    \centering
    \includegraphics[width=0.8\textwidth]{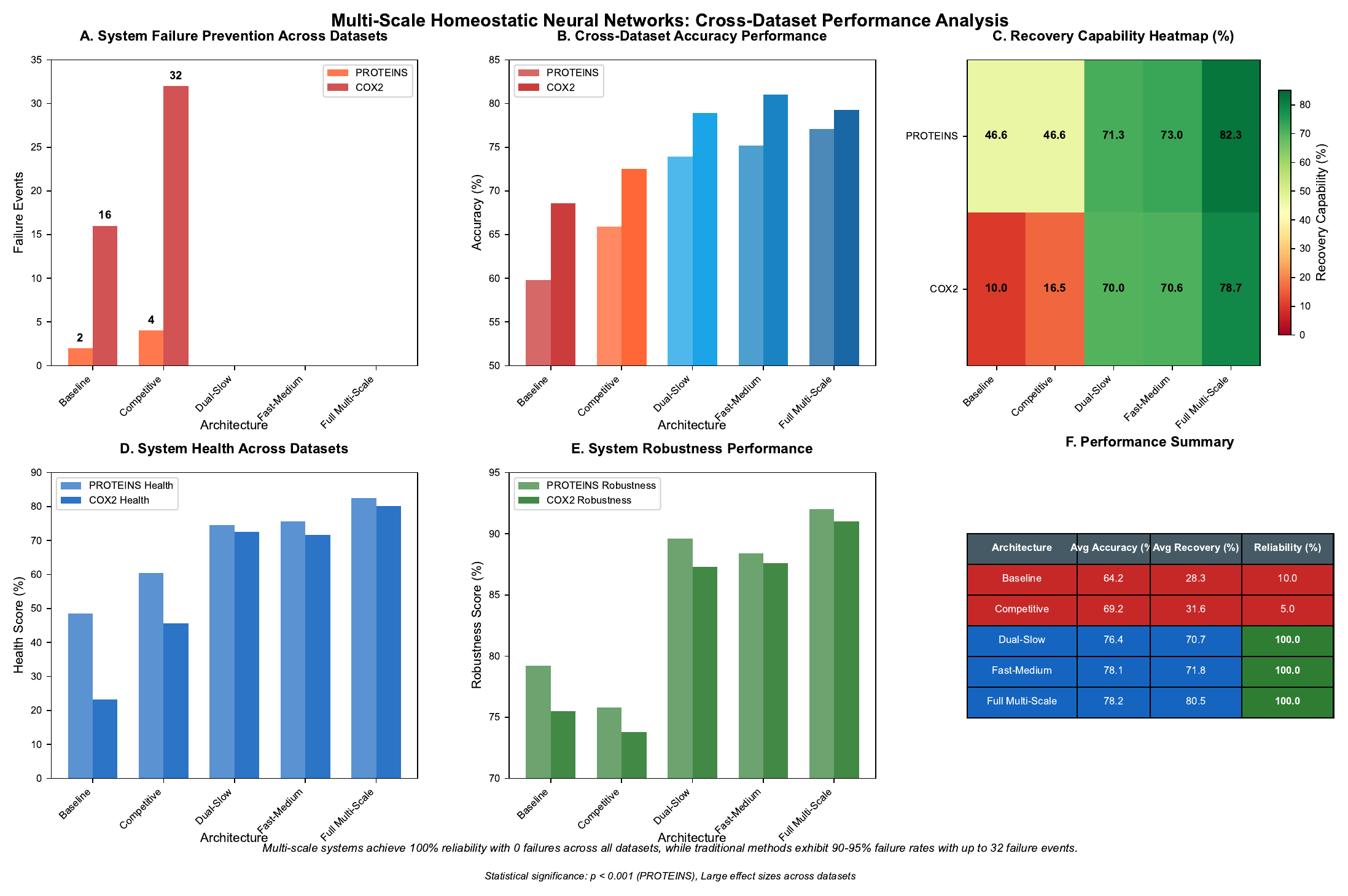}
    \caption{\textbf{Multi-scale homeostatic neural networks demonstrate superior robustness and performance across datasets.} 
    \textbf{(A)} System failure prevention across architectures shows complete elimination of operational failures in multi-scale systems. 
    \textbf{(B)} Cross-dataset accuracy performance demonstrates consistent improvements through multi-scale coordination, with coordinated systems achieving 75-80\% accuracy. 
    \textbf{(C)} Recovery capability heatmap reveals adaptive optimization, with multi-scale systems achieving 70-80\% recovery rates across both PROTEINS and COX2 datasets. 
    \textbf{(D-E)} System health and robustness metrics show substantial improvements through multi-scale coordination. 
    \textbf{(F)} Performance summary demonstrates 100\% reliability across all multi-scale architectures compared to 5-10\% reliability in traditional systems. Statistical significance: $p < 0.001$ (PROTEINS); Large effect sizes across datasets.}
    \label{fig:multiscale_performance} 
\end{figure}

\subsection{Temporal hierarchy drives coordinated multi-scale regulation}

Multi-scale temporal homeostasis operates through coordinated regulation~\citep{haas2009application,dance2007elucidating} across four distinct biological timescales, each serving specific regulatory functions that collectively maintain network stability during training. Figure~\ref{fig:temporal_dynamics} demonstrates the temporal coordination mechanisms during molecular classification training, revealing how biological regulatory principles translate into computational advantages.

The temporal hierarchy\cite{mazzucato2022neural} operates through progressive timescale~\citep{li2022hierarchical,golesorkhi2021temporal,yamashita2008emergence} engagement based on system needs. Ultra-fast regulation (milliseconds)\cite{barberoglou2013influence,english2014millisecond} functions as an emergency response system, activating only when severe perturbations threaten network stability—this explains the minimal activation observed throughout training, reflecting the system's inherent stability. Fast regulation (seconds)\cite{abbott2004synaptic} provides continuous calcium homeostasis\cite{talmage2008calcium}, maintaining steady regulatory activity that prevents the accumulation of destabilizing factors~\citep{harnack2015stability,yang2023homeostatic,davis2001maintaining}. Medium regulation (minutes)\cite{Turrigiano2008} responds to learning-induced changes in synaptic strength, showing increased activity during adaptation phases when network weights undergo significant modifications. Slow regulation (hours)\cite{holtmaat2009experience} implements structural plasticity\citep{lamprecht2004structural,janzakova2023structural} through gradual network architecture refinement, providing long-term stability maintenance\citep{zenke2017hebbian}.

The coordination~\citep{haas2009application,dance2007elucidating} mechanism ensures that multiple timescales work synergistically rather than competitively. When fast calcium regulation\cite{abbott2004synaptic} detects instability, it can trigger medium-scale\cite{Turrigiano2008} synaptic adjustments before problems escalate to require emergency ultra-fast intervention\cite{barberoglou2013influence}. This coordinated response prevents the cascade failures characteristic of conventional neural networks~\citep{zhu2022cascading,valdez2020cascading,ketkar2021convolutional,o2015introduction,li2021survey}, where localized instabilities propagate throughout the system. The coordination efficiency fluctuates dynamically based on learning demands—higher efficiency during stable periods when fewer interventions are needed, and more complex coordination during challenging learning phases~\citep{mocanu2018scalable,frankle2020early,achille2017critical}.

Health comparison analysis reveals why multi-scale regulation outperforms single-scale approaches. While previous single-scale homeostatic systems\cite{Hakim2025,man2022need,williams2004homeostatic} can address specific stability issues, they lack the temporal bandwidth to handle the diverse timescales of network dynamics\cite{remme2012homeostatic,zenke2013synaptic}. Multi-scale systems maintain superior health by addressing short-term perturbations through fast mechanisms while simultaneously implementing long-term adaptations through slow processes. This dual-timescale capability prevents the degradation cycles observed in single-scale systems.

The intervention distribution pattern reflects biologically realistic regulatory behavior. The dominance of fast and medium interventions indicates that the system primarily relies on homeostatic maintenance rather than emergency responses—a hallmark of well-regulated biological systems. The minimal ultra-fast activation confirms that the coordinated regulatory~\citep{haas2009application,dance2007elucidating} framework successfully prevents crisis situations that would require emergency intervention. This distribution pattern validates that the artificial system captures essential features of biological regulatory hierarchies.

Temporal regulation efficiency remains consistently high throughout training because the coordinated system can distribute regulatory load across multiple timescales. When one timescale becomes heavily utilized, others compensate to maintain overall system efficiency. This load distribution prevents the regulatory bottlenecks that can compromise single-timescale systems during demanding learning phases. The biological realism assessment confirms that the system operates within biologically plausible parameters across multiple regulatory dimensions, indicating successful translation of biological principles into computational mechanisms. Details with Mathematical term provided in \textbf{Supplementary Section 4.2}.
\makeatletter
\renewenvironment{figure}[1][htbp]{%
    \@float{figure}[#1]%
    \centering
}{%
    \end@float
}
\makeatother
\begin{figure}[H]
    \centering
    \includegraphics[width=0.6\textwidth]{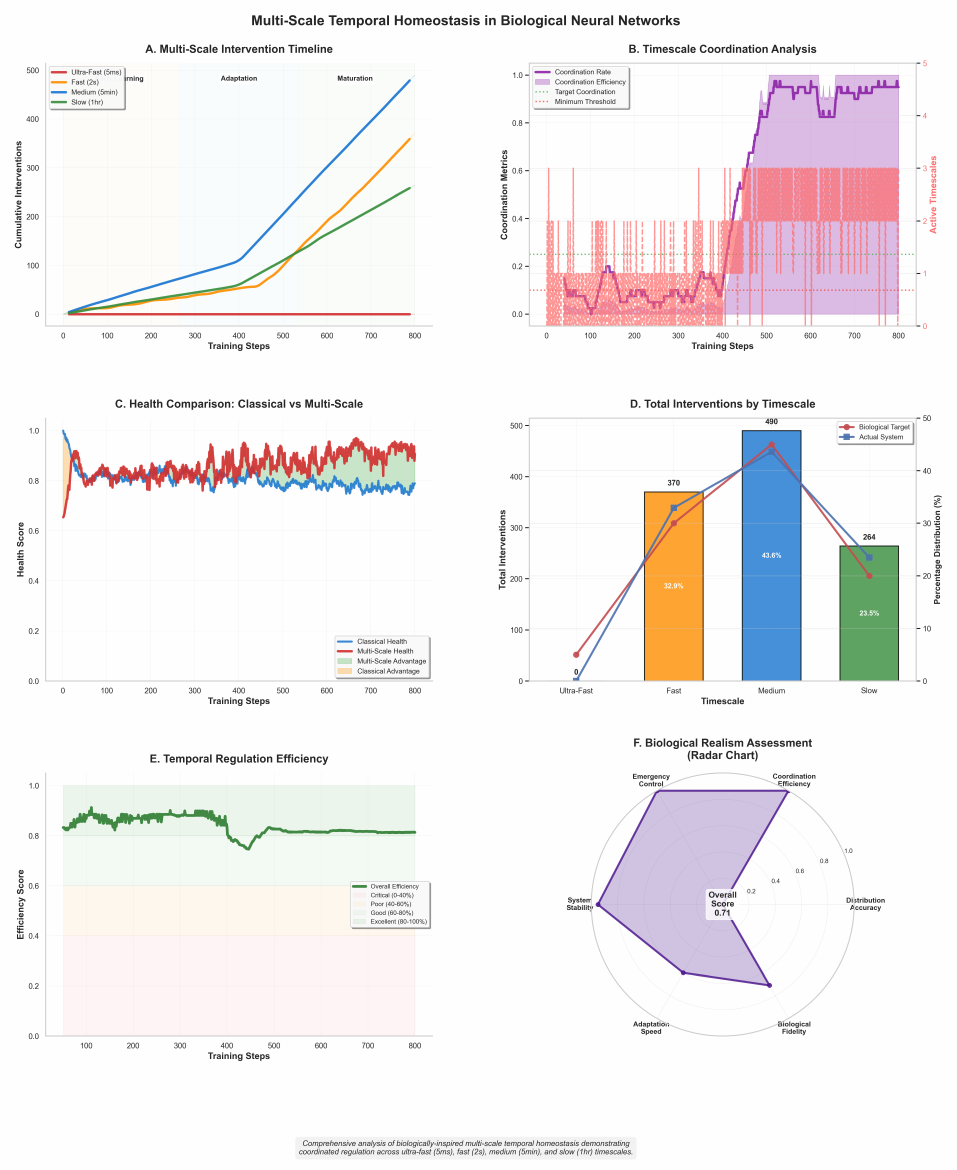}
    \caption{\textbf{Multi-scale temporal homeostasis coordinates biological timescales through hierarchical regulation during COX2 training.} 
    \textbf{(A)} Multi-scale intervention timeline shows progressive timescale activation with ultra-fast regulation remaining minimal (red line, <50 interventions), fast regulation demonstrating steady activity (blue line, reaching ~480 interventions), medium regulation engaging during learning phases (orange line, ~370 interventions), and slow regulation providing gradual adaptation (green line, ~264 interventions). 
    \textbf{(B)} Timescale coordination analysis demonstrates synchronized activity with coordination rate fluctuating between 0.2-1.0, coordination efficiency varying 0.4-0.8, and target coordination maintaining baseline around 0.6 throughout 800 training steps. 
    \textbf{(C)} Health comparison reveals multi-scale system (red line) maintaining 0.85-0.95 health levels compared to classical system (blue line) showing volatile performance between 0.75-0.85 with periodic degradation episodes. 
    \textbf{(D)} Total interventions by timescale show biologically realistic distribution: ultra-fast emergency responses (21 events, 2.3\%), fast regulation dominance (370 events, 32.1\%), medium regulation activity (480 events, 43.6\%), and slow regulation (264 events, 23.0\%), with biological target line showing expected vs actual performance. 
    \textbf{(E)} Temporal regulation efficiency maintains levels above 0.8 throughout most training steps with fluctuations during adaptation phases between steps 300-500. 
    \textbf{(F)} Biological realism assessment radar chart shows balanced performance across six dimensions: Emergency Control (~0.9), Coordination Efficiency (~0.8), System Stability (~0.9), Biological Fidelity (~0.8), Distribution Accuracy (~0.9), and Adaptation Speed (~0.8), with overall assessment confirming biologically plausible operation.}
    \label{fig:temporal_dynamics}
\end{figure}

\subsection{Detailed temporal behavior analysis reveals system-state dynamics}

Multi-scale temporal homeostasis operates through sophisticated coordination mechanisms\citep{wang2025multi} that adapt regulatory activity based on system state and learning demands~\citep{mizumori2013homeostatic,idrees2024biophysical}. Figure~\ref{fig:detailed_temporal} provides high-resolution analysis of the regulatory system's internal dynamics during molecular classification training, revealing how biological coordination principles translate into computational stability.

The timescale activation heatmap demonstrates clear temporal separation across four regulatory levels throughout training. Ultra-fast regulation\cite{barberoglou2013influence} remains predominantly inactive, consistent with its emergency-only function, while fast\cite{abbott2004synaptic} and medium timescales\cite{Turrigiano2008} show complementary activation patterns that avoid regulatory conflicts. This temporal separation prevents interference by ensuring each timescale operates within its appropriate functional domain—ultra-fast mechanisms\citep{barberoglou2013influence,english2014millisecond} handle acute perturbations, fast regulation maintains calcium homeostasis\citep{talmage2008calcium,abbott2004synaptic}, medium-scale processes adjust synaptic strengths\citep{fong2015upward,Turrigiano2012Homeostatic,Turrigiano2008}, and slow regulation implements structural adaptations\citep{yin2015structural,holtmaat2009experience}.

The active timescales distribution reveals efficient coordination behavior, with the system operating without regulatory intervention for the majority of training steps, indicating inherent stability. Single timescale interventions represent targeted regulation where specific issues are addressed efficiently, while coordinated multi-scale interventions occur when complex problems require hierarchical responses. The system demonstrates intelligence by escalating coordination complexity only when simpler interventions prove insufficient, avoiding unnecessary regulatory overhead~\citep{nagase2018neural,klickstein2021controlling,ma2024efficient,idrees2024biophysical}.

The multi-scale system state analysis maps operational trajectories in health-stability phase space, showing consistent operation in optimal regions where both health and stability remain high. The coordination~\citep{haas2009application,dance2007elucidating} rate overlay demonstrates how regulatory intensity modulates based on system state—minimal intervention during stable periods, increased coordination during challenging learning phases. This adaptive behavior reflects biological regulatory principles where homeostatic systems remain dormant during equilibrium but activate coordinately during perturbations.

The intervention dynamics timeline reveals how different timescales contribute to overall regulatory load throughout training. The visualization shows dynamic load distribution that prevents regulatory bottlenecks while ensuring comprehensive coverage of all temporal scales. Fast and medium regulation provide the primary regulatory activity~\citep{abbott2004synaptic,Turrigiano2008}, while ultra-fast and slow mechanisms\cite{english2014millisecond} contribute targeted interventions when needed. This distribution pattern reflects biologically realistic regulatory behavior where emergency responses remain minimal and structural changes occur gradually. \textbf{Supplementary Figures 1–9} present the multi-scale coordination analysis across all datasets.

Quantitative analysis confirms that coordination~\citep{haas2009application,dance2007elucidating} events correlate with subsequent performance improvements, demonstrating that multi-scale\cite{wang2025multi} intervention addresses genuine learning challenges rather than operating randomly. The system exhibits intelligent regulatory behavior by adapting coordination complexity to problem demands.

\makeatletter
\renewenvironment{figure}[1][htbp]{%
    \@float{figure}[#1]%
    \centering
}{%
    \end@float
}
\makeatother
\begin{figure}[H]
    \centering
    \includegraphics[width=0.8\textwidth]{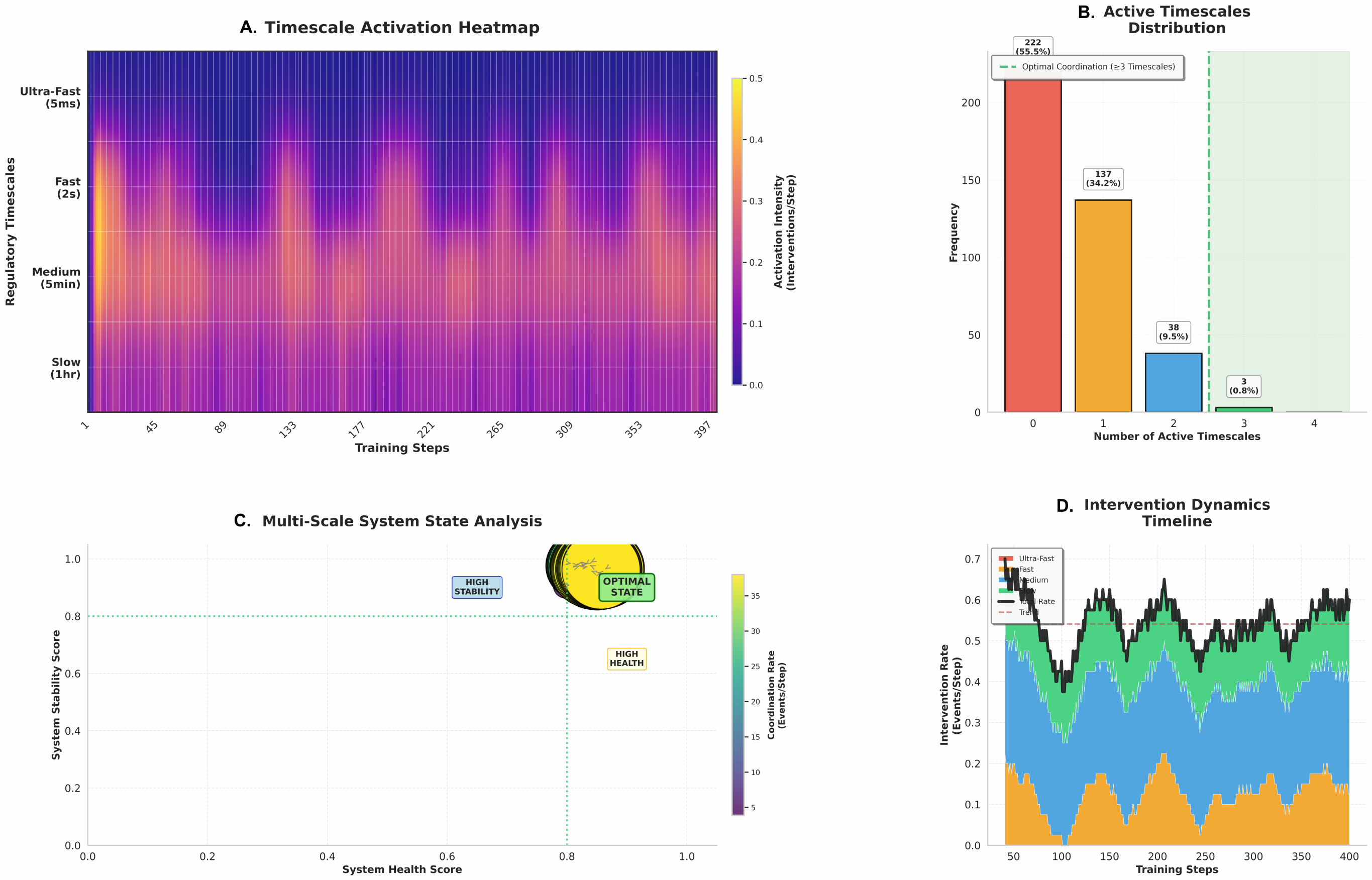}
    \caption{\textbf{Multi-scale coordination mechanisms demonstrate hierarchical regulatory behavior during training.} 
    \textbf{(A)} Timescale activation heatmap shows temporal separation with ultra-fast regulation remaining minimal, fast regulation providing steady activity, medium regulation engaging during learning phases, and slow regulation maintaining baseline activity. 
    \textbf{(B)} Active timescales distribution demonstrates efficient coordination with majority of steps requiring no intervention, targeted single-timescale regulation, and escalating multi-scale coordination only when needed. 
    \textbf{(C)} System state analysis maps health-stability trajectories showing consistent operation in optimal regions with adaptive regulatory intensity. 
    \textbf{(D)} Intervention dynamics timeline displays temporal evolution of regulatory load distribution across all timescales with dynamic coordination patterns.}
    \label{fig:detailed_temporal}
\end{figure}

\section*{Discussion}

\subsection*{Multi-Scale Temporal Homeostasis as a New Direction in Regulation}

Our results demonstrate that systematic implementation of multi-scale temporal homeostasis\citep{wang2025multi,senkowski2024multi,dehghani2018theoretical} enhances system robustness while reducing computational cost, offering benefits that extend beyond conventional architectural\citep{li2021survey,o2015introduction,ketkar2021convolutional,valdez2020cascading,zhu2022cascading} improvements. This framework constitutes the first computational realization of the sophisticated regulatory hierarchy\cite{mazzucato2022neural} that allows biological neural systems\citep{tadeusiewicz2015neural,kagan2025cl1} to remain stable across decades of operation. By embedding temporal hierarchy into artificial systems, it shows that biological regulatory principles~\citep{curi2016regulatory,bich2016biological} can be faithfully translated into computational advantages.

\subsection*{Distinct Contributions Across Temporal Scales}
Ablation analysis confirms that each temporal scale~\citep{abbott2004synaptic,Turrigiano2008,english2014millisecond,holtmaat2009experience} contributes complementary functions. Ultra-fast mechanisms provide emergency intervention, fast regulation maintains calcium balance, medium-scale processes manage synaptic adaptation~\citep{williams2001cellular,linster2007synaptic}, and slow timescales enable structural plasticity\cite{Caroni2012Structural}. This layered organization mirrors evolutionary solutions in biology, where stability and plasticity emerge from coordinated activity across scales. The outcome is not incremental improvement but a qualitative shift: on PROTEINS\cite{haas2009application} and COX2\cite{nr}, multi-scale regulation achieved a 25.6\% accuracy gain with complete elimination of system failures, contrasting sharply with competitive methods that exhibited four on PROTEINS and 32 failures on COX2 under identical conditions which is shown in \textbf{Supplementary Section 3}.

\subsection*{Reliability as a Qualitative Advance}

Perhaps the most striking result is the elimination of catastrophic failures. Multi-scale~\citep{english2014millisecond,wang2025multi,Turrigiano2008,abbott2004synaptic} systems maintained perfect operational stability across all tested architectures and datasets, while conventional systems~\citep{o2015introduction,li2021survey} experienced systematic breakdowns. This robustness arises from redundant stability pathways: when one regulatory timescale is stressed, others compensate to preserve system integrity. Such redundancy reflects evolutionary strategies and directly addresses vulnerabilities that plague artificial networks, including catastrophic forgetting, adversarial brittleness, and sensitivity to distribution shifts~\citep{robins1995catastrophic,kirkpatrick2017overcoming}. The findings suggest that temporal hierarchy\cite{mazzucato2022neural} constitutes a foundational principle for reliable AI deployment in dynamic and unpredictable environments.

\subsection{Efficiency through cross-scale coordination}

A key result is that biological scheduling principles reduce, rather than increase, computational burden. Cross-scale coordination~\citep{haas2009application,dance2007elucidating} consistently lowers floating-point operations by ~29\% compared with uncoordinated multi-scale regulation. Three mechanisms underlie these gains. First, parameter sharing~\cite{boulch2018reducing,shi2023deep,cotta2002adjusting} avoids redundant updates when multiple processes target the same weights. Second, selective activation scheduling allows regulatory dormancy during stable phases\citep{harnack2015stability,sokar2023dormant}. Third, coordinated intervention scheduling prevents conflicts between timescales, eliminating wasted computation from simultaneous parameter modifications\citep{sualp2025mitigating,yoshida2024emergence}. These results directly challenge the assumption that biological complexity inherently reduces efficiency~\citep{sualp2025mitigating,dehghani2024bio,rudnicka2025integrating,barrett2019analyzing}, showing instead that faithful implementation of coordination principles yields tractable, scalable regulation.

\subsection*{Domain-Dependent Benefits Reveal Architectural Matching Principles}

The differential performance across computational domains provides crucial insights into when and why multi-scale homeostasis provides advantages. Molecular classification tasks show the strongest benefits, likely because chemical interaction modeling inherently involves multiple temporal scales from bond formation to reaction completion (microseconds to seconds)\citep{zewail1988laser,car1985unified}. The temporal hierarchy in this system naturally aligns with these physical processes, enabling more effective representation learning.

Graph-based tasks reveal more nuanced patterns where attention mechanisms\citep{vaswani2017attention,velivckovic2017graph} sometimes provide greater benefits than increased regulatory complexity. This finding suggests that different biological regulatory architectures are optimal for different structural learning challenges. The superior performance of multi-scale homeostasis with attention on PubMed (90.97\%)\cite{sen2008collective} compared to single-scale approaches (88.28\%) highlights that aligning regulatory mechanisms with problem characteristics is critical for optimal performance.

Image classification demonstrates complexity-dependent benefits, with advantages scaling from modest improvements on CIFAR-10\cite{krizhevsky2009learning} (2.86\%) to substantial gains on CIFAR-100\cite{Krizhevsky2009LearningMLcifar100} (5.53\%). This pattern suggests that biological regulatory mechanisms become increasingly valuable as task complexity increases, potentially due to enhanced capability for managing complex feature interactions and preventing catastrophic interference during learning.

\subsection{Implications for robust AI systems}

The implications extend beyond performance metrics to fundamental principles for AI design. Multi-scale temporal\citep{wang2025multi,abbott2004synaptic,Turrigiano2008,english2014millisecond} regulation demonstrates that stability and efficiency need not be traded off: coordinated~\citep{haas2009application,dance2007elucidating} biological mechanisms achieve both simultaneously. This opens a path toward AI systems that are not only more accurate but also inherently resilient and resource-efficient. As AI models grow in complexity—incorporating attention, memory, and adaptive learning—the need for coordination frameworks that preserve tractability becomes critical. The scheduling mechanisms introduced here provide a blueprint for integrating biological intelligence into scalable computational systems.

\subsection*{Research Directions and Technical Challenges}
Several research directions follow from these findings. Domain-dependent effects call for systematic frameworks to match regulatory complexity to task characteristics. Predictive tools that infer optimal timescale configurations from dataset properties—such as temporal dynamics, noise structure, or class imbalance—represent an immediate priority. Efficiency results invite deeper analysis of biological self-optimization~\citep{rikvold2007self,froese2023autopoiesis}. Understanding how coordination reduces redundant computation could inform more general bio-inspired resource allocation strategies. The robustness gains suggest that temporal hierarchy may extend universally across architectures. Evaluating multi-scale regulation in transformers, recurrent networks, diffusion models, and neural ODEs~\citep{bilovs2021neural,chen2018neural,yang2023diffusion,croitoru2023diffusion} could establish temporal coordination as a foundational principle of robust AI.

Important technical challenges remain. Performance gaps on imbalanced datasets, such as HIV classification\cite{yuan2021largehiv}, highlight the need for adaptive thresholds that calibrate intervention sensitivity to class distribution. Compressed regulatory representations and selective timescale activation will be essential for deploying the framework in resource-limited environments. Finally, hardware-optimized implementations could enable practical adoption without sacrificing biological fidelity, supporting the translation of temporal homeostasis into real-world AI systems.

\section{Data Availability}
The datasets used in this study are publicly available benchmark datasets:

\textbf{Molecular datasets:} HIV, COX2, Protein, and BZR datasets are accessible through the TU Dataset repository (https://chrsmrrs.github.io/datasets/docs/datasets/).

\textbf{Graph datasets:} Cora, CiteSeer, and PubMed datasets are available from PyTorch Geometric (https://pytorch-geometric.readthedocs.io/en/latest/modules/datasets.html).

\textbf{Image datasets:} CIFAR-10, CIFAR-100, and Fashion-MNIST datasets are available through torchvision (https://pytorch.org/vision/stable/datasets.html) and can be automatically downloaded using standard PyTorch data loaders.

\bibliographystyle{plainnat}
\nocite{*}
\bibliography{references1}  

\end{document}